\documentclass[11pt,a4paper]{article}
\usepackage[hyperref]{naaclhlt2018}
\usepackage{times}
\usepackage{latexsym}

\usepackage{url}
\usepackage{algorithm}
\usepackage{algpseudocode}

\usepackage{graphicx}

\usepackage{amsmath}

\usepackage{booktabs}

\usepackage{multirow}

\usepackage{tikz}

\usepackage{pgfplots}

\usepackage{caption}

\usepackage{pifont}

\usepackage[T1]{fontenc} 

\usepackage{subfigure}
\aclfinalcopy % Uncomment this line for the final submission
\title{Improving Character-based Decoding Using Target-Side Morphological Information for Neural Machine Translation}

\author{Peyman Passban, Qun Liu, Andy Way\\
   ADAPT Centre\\
   School of Computing\\
   Dublin City University, Ireland\\
   {\tt firstname.lastname@adaptcentre.ie}
}

\date{}

\begin{document}
\maketitle

\begin{abstract}
Recently, neural machine translation (NMT) has emerged as a powerful alternative to conventional statistical approaches. However, its performance drops considerably in the presence of morphologically rich languages (MRLs). Neural engines usually fail to tackle the large vocabulary and high out-of-vocabulary (OOV) word rate of MRLs. Therefore, it is not suitable to exploit existing word-based models to translate this set of languages. In this paper, we propose an extension to the state-of-the-art model of \newcite{chung-cho-bengio}, which works at the character level and boosts the decoder with target-side morphological information. In our architecture, an additional morphology table is plugged into the model. Each time the decoder samples from a target vocabulary, the table sends auxiliary signals from the most relevant affixes in order to enrich the decoder's current state and constrain it to provide better predictions. We evaluated our model to translate English into German, Russian, and Turkish as three MRLs and observed significant improvements.
\end{abstract}

\section{Introduction}
Morphologically complex words (MCWs) are multi-layer structures which consist of different subunits, each of which carries semantic information and has a specific syntactic role. Table \ref{mcw} gives a Turkish example to show this type of complexity. This example is a clear indication that word-based models are not suitable to process such complex languages. Accordingly, when translating MRLs, it might not be a good idea to treat words as atomic units as it demands a large vocabulary that imposes extra overhead. Since MCWs can appear in various forms we require a very large vocabulary to $i$) cover as many morphological forms and words as we can, and $ii$) reduce the number of OOVs. Neural models by their nature are complex, and we do not want to make them more complicated by working with large vocabularies. Furthermore, even if we have quite a large vocabulary set, clearly some words would remain uncovered by that. This means that a large vocabulary not only complicates the entire process, but also does not necessarily mitigate the OOV problem. For these reasons we propose an NMT engine which works at the character level. 
\begin{table}[th]
	\centering
	\small
	\begin{tabular}{l l}
		%\toprule
		{\bf Word} & {\bf Translation}\\
		\midrule
		{\bf terbiye} & good manners \\
        {\bf terbiye}\textbf{.}siz & rude \\
        {\bf terbiye}\textbf{.}siz\textbf{.}lik & rudeness\\
        {\bf terbiye}\textbf{.}siz\textbf{.}lik\textbf{.}leri & their rudeness\\
        {\bf terbiye}\textbf{.}siz\textbf{.}lik\textbf{.}leri\textbf{.}nden & from their rudeness \\ 
        %{\bf terbiye}+siz+lik+leri+nden+mis & it was because of their rudeness\\
		%\bottomrule
	\end{tabular}
	\caption{\label{mcw}\small Illustrating subword units in MCWs. The boldfaced part indicates the stem.}
\end{table} 

In this paper, we focus on translating into MRLs and issues associated with word formation on the target side. To provide a better translation we do not necessarily need a large target lexicon, as an MCW can be gradually formed during decoding by means of its subunits, similar to the solution proposed in character-based decoding models \cite{chung-cho-bengio}. Generating a complex word character-by-character is a better approach compared to word-level sampling, but it has other disadvantages. 

One character can co-occur with another with almost no constraint, but a particular word or morpheme can only collocate with a very limited number of other constituents.  Unlike words, characters are not meaning-bearing units and do not preserve syntactic information, so (in the extreme case) the chance of sampling each character by the decoder is almost equal to the others, but this situation is less likely for words. The only constraint that prioritize which character should be sampled is information stored in the decoder, which we believe is insufficient to cope with all ambiguities. Furthermore, when everything is segmented into characters the target sentence with a limited number of words is changed to a very long sequence of characters, which clearly makes it harder for the decoder to remember such a long history. Accordingly, character-based information flows in the decoder may not be as informative as word- or morpheme-based information.

In the character-based NMT model everything is almost the same as its word-based counterpart except the target vocabulary whose size is considerably reduced from thousands of words to just hundreds of characters. If we consider the decoder as a classifier, it should in principle be able to perform much better over hundreds of classes (characters) rather than thousands (words), but the performance of character-based models is almost the same as or slightly better than their word-based versions. This underlines the fact that the character-based decoder is perhaps not fed with sufficient information to provide improved performance compared to word-based models.

Character-level decoding limits the search space by dramatically reducing the size of the target vocabulary, but at the same time widens the search space by working with characters whose sampling seems to be harder than words. The freedom in selection and sampling of characters can mislead the decoder, which prevents us from taking the maximum advantages of character-level decoding. If we can control the selection process with other constraints, we may obtain further benefit from restricting the vocabulary set, which is the main goal followed in this paper.    

In order to address the aforementioned problems we redesign the neural decoder in three different scenarios. In the first scenario we equip the decoder with an additional morphology table including target-side affixes. We place an attention module on top of the table which is controlled by the decoder. At each step, as the decoder samples a character, it searches the table to find the most relevant information which can enrich its state. Signals sent from the table can be interpreted as additional constraints. In the second scenario we share the decoder between two output channels. The first one samples the target character and the other one predicts the morphological annotation of the character. This multi-tasking approach forces the decoder to send morphology-aware information to the final layer which results in better predictions. In the third scenario we combine these two models. Section \ref{models} provides more details on our models. 

Together with different findings that will be discussed in the next sections, there are two main contributions in this paper. We redesigned and tuned the NMT framework for translating into MRLs. It is quite challenging to show the impact of external knowledge such as morphological information in neural models especially in the presence of large parallel corpora. However, our models are able to incorporate morphological information into decoding and boost its quality. We inject the decoder with morphological properties of the target language. Furthermore, the novel architecture proposed here is not limited to morphological information alone and is flexible enough to provide other types of information for the decoder. 

\section{NMT for MRLs}\label{nmt-review}
There are several models for NMT of MRLs which are designed to deal with morphological complexities. \newcite{garcia2016factored} and \newcite{sennrich-haddow:2016:WMT} adapted the factored machine translation approach to neural models. Morphological annotations can be treated as extra factors in such models. \newcite{jean-EtAl:2015:ACL-IJCNLP} proposed a model to handle very large vocabularies. \newcite{luong-EtAl:2015:ACL-IJCNLP} addressed the problem of rare words and OOVs with the help of a post-translation phase to exchange unknown tokens with their potential translations.  \newcite{sennrich2015neural} used subword units for NMT. The model relies on frequent subword units instead of words. \newcite{costajussa-fonollosa:2016:P16-2} designed a model for translating from MRLs. The model encodes source words with a convolutional module proposed by \newcite{kim2015character}. Each word is represented by a convolutional combination of its characters. 

\newcite{luong-manning:2016:P16-1} used a hybrid model for representing words. In their model, unseen and complex words are encoded with a character-based representation, with other words encoded via the usual surface-form embeddings. \newcite{DBLP:journals/corr/VylomovaCHH16} compared different representation models (word-, morpheme, and character-level models) which try to capture complexities on the source side, for the task of translating from MRLs.

\newcite{chung-cho-bengio} proposed an architecture which benefits from different segmentation schemes. On the encoder side, words are segmented into subunits with the {byte-pair} segmentation model (\textit{bpe}) \cite{sennrich2015neural}, and on the decoder side, one target character is produced at each time step. Accordingly, the target sequence is treated as a long chain of characters without explicit segmentation. \newcite{W17-4727} focused on translating from English into Finnish and  implicitly incorporated morphological information into NMT through multi-task learning.  \newcite{passbanPhD} comprehensively studied the problem of translating MRLs and addressed potential challenges in the field.

Among all the models reviewed in this section, the network proposed by \newcite{chung-cho-bengio} could be seen as the best alternative for translating into MRLs as it works at the character level on the decoder side and it was evaluated in different settings on different languages. Consequently, we consider it as a baseline model in our experiments.

\section{Proposed Architecture}\label{models}
We propose a compatible neural architecture for translating into MRLs. The model benefits from subword- and character-level information and improves upon the state-of-the-art model of \newcite{chung-cho-bengio}. We manipulated the model to incorporate morphological information and developed three new extensions, which are discussed in Sections \ref{embed-tab}, \ref{multi-task}, and \ref{both}. 

\subsection{The Embedded Morphology Table}\label{embed-tab}
In the first extension an additional table containing the morphological information of the target language is plugged into the decoder to assist with word formation. Each time the decoder samples from the target vocabulary, it searches the morphology table to find the most relevant affixes given its current state. Items selected from the table act as guiding signals to help the decoder sample a better character. 

Our base model is an encoder-decoder model with attention \cite{bahdanau2014neural}, implemented using gated recurrent units (GRUs) \cite{cho-EtAl:2014:EMNLP2014}. We use a four-layer model in our experiments. Similar to \newcite{chung-cho-bengio} and \newcite{DBLP:journals/corr/WuSCLNMKCGMKSJL16}, we use bidirectional units to encode the source sequence. Bidirectional GRUs are placed only at the input layer. The forward GRU reads the input sequence in its original order and the backward GRU reads the input in the reverse order. Each hidden state of the encoder in one time step is a concatenation of the forward and backward states at the same time step. This type of bidirectional processing provides a richer representation of the input sequence. 

On the decoder side, one target character is sampled from a target vocabulary at each time step. In the original encoder-decoder model, the probability of predicting the next token $y_i$ is estimated based on $i$) the current hidden state of the decoder, $ii$) the last predicted token, and $iii$) the context vector. This process can be formulated as $p(y_i|y_1,...,y_{i-1},{\bf x}) = g(h_i,y_{i-1},{\bf c}_i)$, where $g(.)$ is a softmax function, $y_i$ is the target token (to be predicted), $\textbf{x}$ is the representation of the input sequence, $h_i$ is the decoder's hidden state at the $i$-th time step, and ${\bf c}_i$ indicates the context vector which is a weighted summary of the input sequence generated by the attention module. ${\bf c}_i$ is generated via the procedure shown in \eqref{alphavals}:
\begin{equation}
\label{alphavals}
\begin{aligned}
{\bf c}_i &= \sum_{j=1}^{n} \alpha_{ij} s_j\\
\alpha_{ij} &=\frac{\exp{(e_{ij})}}{\sum{_{k=1}^{n}\exp{(e_{ik})}}}; \hspace{2mm}e_{ij}=a(s_j, h_{i-1})
\end{aligned}
\end{equation}
where $\alpha_{ij}$ denotes the weight of the $j$-th hidden state of the encoder ($s_j$) when the decoder predicts the $i$-th target token, and $a()$ shows a combinatorial function which can be modeled through a simple feed-forward connection. $n$ is the length of the input sequence.
\begin{figure*}[ht]
    \centering
    \includegraphics{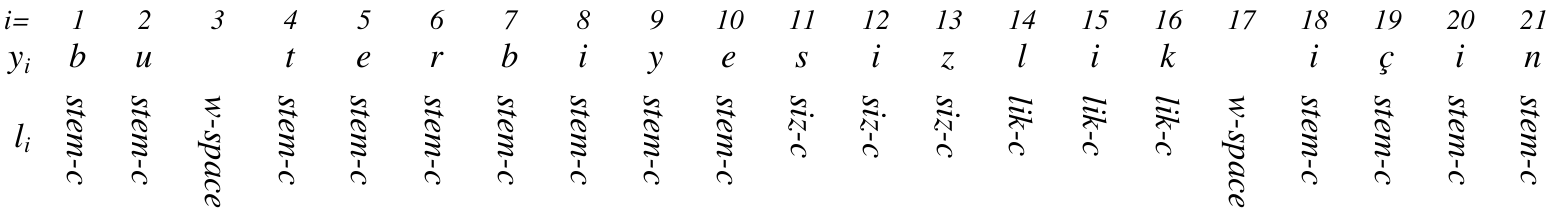}
    \caption{\label{tags}The target label that each output channel is supposed to predict when generating the Turkish sequence `\textit{bu$_1$ terbiyesizlik$_2$ i\c{c}in$_3$}' meaning `\textit{because$_3$ of$_3$ this$_1$ rudeness$_2$}'.}
\end{figure*}

In our first extension, the prediction probability is conditioned on one more constraint in addition to those three existing ones, as in $p(y_i|y_1,...,y_{i-1},{\bf x}) = g(h_i,y_{i-1},{\bf c}_i, {\bf c}^m_i)$, where ${\bf c}^m_i$ is the morphological context vector and carries information from those useful affixes which can enrich the decoder's information. ${\bf c}^m_i$ is generated via an attention module over the morphology table which works in a similar manner to word-based attention model. The attention procedure for generating ${\bf c}^m_i$ is formulated as in \eqref{aff-atten}:
\begin{equation}
\label{aff-atten}
\begin{aligned}
{\bf c}^m_i &= \sum_{u=1}^{|\mathcal{A}|} \beta_{iu} f_u\\
\beta_{iu} &=  \frac{\exp{(e^m_{iu})}}{\sum{_{v=1}^{|\mathcal{A}|} \exp{(e_{iv})}}}; \hspace{2mm}e^m_{iu}= a^m(f_u, h_{i-1})
\end{aligned}
\end{equation}
where $f_u$ represents the embedding of the $u$-th affix ($u$-th column) in the morphology/affix table $\mathcal{A}$, $\beta_{iu}$ is the weight assigned to $f_u$ when predicting the $i$-th target token, and $a^m$ is a feed-forward connection between the morphology table and the decoder.  

The attention module in general can be considered as a search mechanism, e.g. in the original encoder-decoder architecture the basic attention module finds the most relevant input words to make the prediction. In multi-modal NMT \cite{huang2016attention,calixto-liu-campbell:2017:Long} an extra attention module is added to the basic one in order to search the image input to find the most relevant image segments. In our case we have a similar additional attention module which searches the morphology table. 

In this scenario, the morphology table including the target language's affixes can be considered as an external knowledge repository that sends auxiliary signals which accompany the main input sequence at all time steps. Such a table certainly includes useful information for the decoder. As we are not sure which affix preserves those pieces of useful information, we use an attention module to search for the best match. The attention module over the table works as a filter which excludes irrelevant affixes and amplifies the impact of relevant ones by assigning different weights ($\beta$ values). 

\subsection{The Auxiliary Output Channel}\label{multi-task}
In the first scenario, we embedded a morphology table into the decoder in the hope that it can enrich sampling information. Mathematically speaking, such an architecture establishes an extra constraint for sampling and can control the decoder's predictions. However, this is not the only way of constraining the decoder. In the second scenario, we define extra supervision to the network via another predictor (output channel). The first channel is responsible for generating translations and predicts one character at each time step, and the other one tries to understand the morphological status of the decoder by predicting the morphological annotation ($l_i$) of the target character. 

The approach in the second scenario proposes a multi-task learning architecture, by which in one task we learn translations and in the other one morphological annotations. Therefore, all network modules --especially the last hidden layer just before the predictors-- should provide information which is useful enough to make correct predictions in both  channels, i.e. the decoder should preserve translation as well as morphological knowledge. Since we are translating into MRLs this type of mixed information (morphology+translation) can be quite useful. 

In our setting, the morphological annotation $l_i$ predicted via the second channel shows to which part of the word or morpheme the target character belongs, i.e. the label for the character is the morpheme that includes it. We clarify the prediction procedure via an example from our training set (see Section \ref{exp}). When the Turkish word `\textit{terbiyesizlik}' is generated, the first channel is supposed to predict \textit{t}, \textit{e}, \textit{r}, up to \textit{k}, one after another. For the same word, the second channel is supposed to predict \textit{stem-C} for the fist $7$ steps as the first $7$ characters `\textit{terbiye}' belong to the stem of the word. The \textit{C} sign indicates that \textit{stem-C} is a class label. The second channel should also predict \textit{siz-C} when the first channel predicts \textit{s} (eighth character), \textit{i} (ninth character), and \textit{z} (tenth character), and \textit{lik-C} when the first channel samples the last three characters. Clearly, the second channel is a classifier which works over the \{\textit{stem-C}, \textit{siz-C}, \textit{lik-C}, ...\} classes. Figure \ref{tags} illustrates a segment of a sentence including this Turkish word and explains which class tags should be predicted by each channel. 

To implement the second scenario we require a single-source double-target training corpus: [source sentence] $\rightarrow$ [sequence of target characters $\&$ sequence of morphological annotations] (see Section \ref{exp}). The objective function should also be manipulated accordingly. Given a training set $\{{\bf x}_t, {\bf y}_t, {\bf m}_t\}_{t=1}^{T}$ the goal is to maximize the joint loss function shown in \eqref{joint}: 
\begin{equation}\label{joint}
\lambda\sum_{t=1}^{T}\log{P({\bf y}_t|{\bf x}_t;\theta)} + (1-\lambda) \sum_{t=1}^{T}\log{P({\bf m}_t|{\bf x}_t;\theta)}
\end{equation}
where $\textbf{x}_t$ is the $t$-th input sentence whose translation is a sequence of target characters shown by $\textbf{y}_t$. $\textbf{m}_t$ is the sequence of morphological annotations and $T$ is the size of the training set. $\theta$ is the set of network parameters and $\lambda$ is a scalar to balance the contribution of each cost function. $\lambda$ is adjusted on the development set during training. 

\subsection{Combining the Extended Output Layer and the Embedded Morphology Table}\label{both}
In the first scenario, we aim to provide the decoder with useful information about morphological properties of the target language, but we are not sure whether signals sent from the table are what we really need. They might be helpful or even harmful, so there should be a mechanism to control their quality. In the second scenario we also have a similar problem as the last layer requires some information to predict the correct morphological class through the second channel, but there is no guarantee to ensure that information in the decoder is sufficient for this sort of prediction. In order to address these problems, in the third extension we combine both scenarios as they are complementary and can potentially help each other.  

The morphology table acts as an additional useful source of knowledge as it already consists of affixes, but its content should be adapted according to the decoder and its actual needs. Accordingly, we need a trainer to update the table properly. The extra prediction channel plays this role for us as it forces the network to predict the target language's affixes at the output layer. The error computed in the second channel is back-propagated to the network including the morphology table and updates its affix information into what the decoder actually needs for its prediction. Therefore, the second output channel helps us train better affix embeddings. 

\begin{figure}[h]
    \centering
    \includegraphics[width=0.45\textwidth]{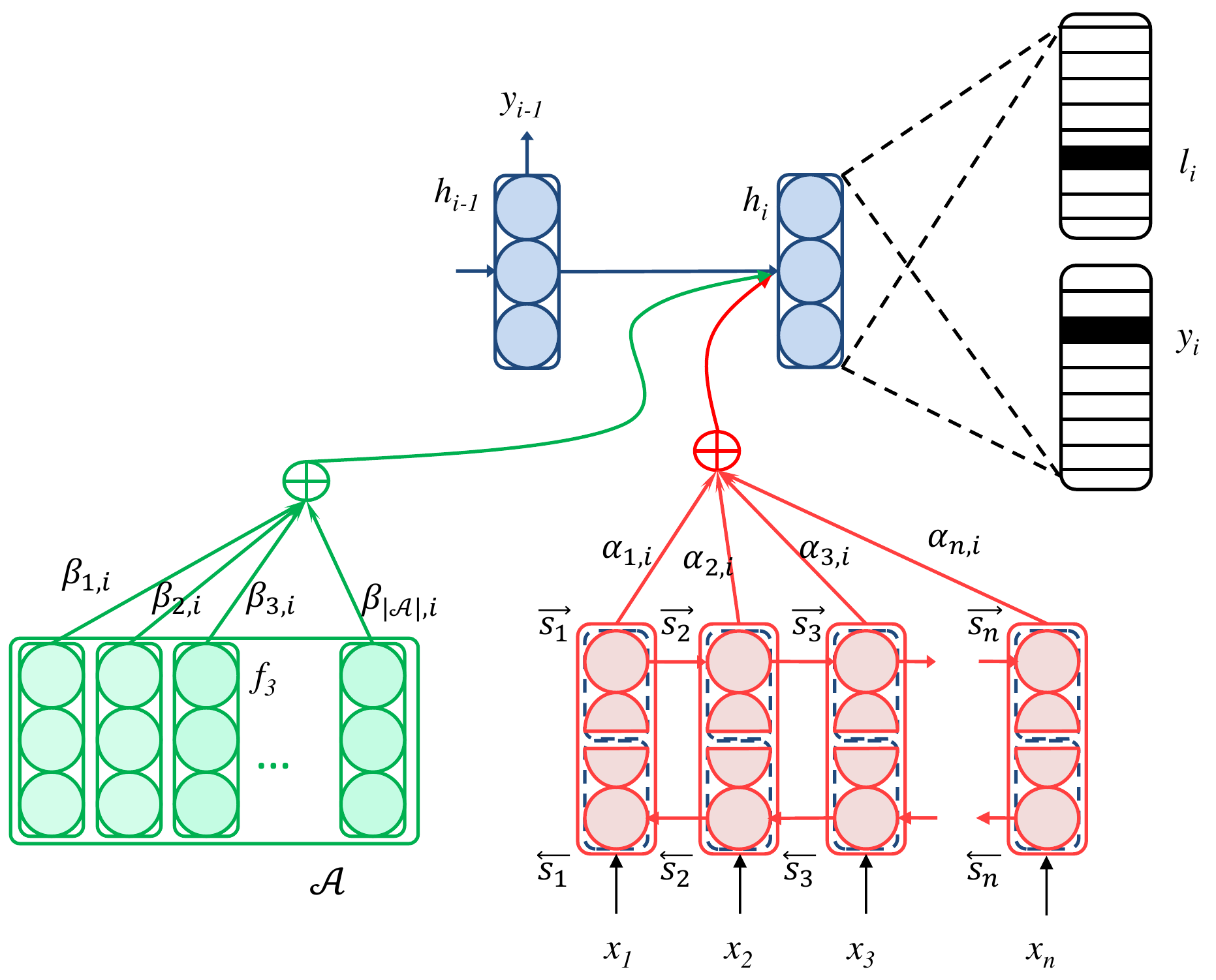}
    \caption{\label{net-arch}The architecture of the NMT model with an auxiliary prediction channel and an extra morphology table. This network includes only one decoder layer and one encoder layer. $\oplus$ shows the attention modules.}
\end{figure}

The morphology table also helps the second predictor. Without considering the table, the last layer only includes information about the input sequence and previously predicted outputs, which is not directly related to morphological information. The second attention module retrieves useful affixes from the morphology table and concatenates to the last layer, which means the decoder is explicitly fed with morphological information. Therefore, these two modules mutually help each other. The external channel helps update the morphology table with high-quality affixes (backward pass) and the table sends its high-quality signals to the prediction layer (forward pass). The relation between these modules and the NMT architecture is illustrated in Figure \ref{net-arch}.
\section{Experimental Study}\label{exp}
As previously reviewed, different models try to capture complexities on the encoder side, but to the best of our knowledge the only model which proposes a technique to deal with complex constituents on the decoder side is that of \newcite{chung-cho-bengio}, which should be an appropriate baseline for our comparisons. Moreover, it outperforms other existing NMT models, so we prefer to compare our network to the best existing model. This model is referred to as CDNMT in our experiments. In the next sections first we explain our experimental setting, corpora, and how we build the morphology table (Section \ref{exp1}), and then report experimental results (Section \ref{exp2}).  

\subsection{Experimental Setting}\label{exp1}
In order to make our work comparable we try to follow the same experimental setting used in CDNMT, where the GRU size is $1024$, the affix and word embedding size is $512$, and the beam width is $20$. Our models are trained using stochastic gradient descent with Adam \citep{adam}. \newcite{chung-cho-bengio} and \newcite{sennrich2015neural} demonstrated that {\it bpe} boosts NMT, so similar to CDNMT we also preprocess the source side of our corpora using {\it bpe}. We use {\tt WMT-15} corpora\footnote{\url{http://www.statmt.org/wmt15/}} to train the models, {\tt newstest-2013} for tuning and {\tt newstest-2015} as the test sets. For English--Turkish (En--Tr) we use the {\tt OpenSubtitle2016} collection \citep{lison2016opensubtitles2016}. The training side of the English--German (En--De), English--Russian (En--Ru), and En--Tr corpora include $4.5$, $2.1$, and $4$ million parallel sentences, respectively. We randomly select $3$K sentences for each of the development and test sets for En--Tr. For all language pairs we keep the $400$ most frequent characters as the target-side character set and replace the remainder (infrequent characters) with a specific character. 

One of the key modules in our architecture is the morphology table. In order to implement it we use a look-up table whose columns include embeddings for the target language's affixes (each column represents one affix) which are updated during training. As previously mentioned, the table is intended to provide useful, morphological information so it should be initialized properly, for which we use a morphology-aware embedding-learning model. To this end, we use the neural language model of \newcite{botha2014compositional} in which each word is represented via a linear combination of the embeddings of its surface form and subunits, e.g. $\overrightarrow{terbiyesizlik} = \overrightarrow{terbiyesizlik} + \overrightarrow{terbiye} + \overrightarrow{siz} + \overrightarrow{lik}$. Given a sequence of words, the neural language model tries to predict the next word, so it learns sentence-level dependencies as well as intra-word relations. The model trains surface form and subword-level embeddings which provides us with high-quality affix embeddings. 

Our neural language model is a recurrent network with a single $1000$-dimensional GRU layer, which is trained on the target sides of our parallel corpora. The embedding size is $512$ and we use a batch size of $100$ to train the model. Before training the neural language model, we need to manipulate the training corpus to decompose words into morphemes for which we use Morfessor \cite{smit-EtAl:2014:Demos}, an unsupervised morphological analyzer. Using Morfessor each word is segmented into different subunits where we consider the longest part as the stem of each word; what appears before the stem is taken as a member of the set of prefixes (there might be one or more prefixes) and what follows the stem is considered as a member of the set of suffixes.

Since Morfessor is an unsupervised analyzer, in order to minimize segmentation errors and avoid noisy results we filter its output and exclude subunits which occur fewer than $500$ times.\footnote{The number may seem a little high, but for a corpus with more than $115$M words this is not a strict threshold in practice.} After decomposing, filtering, and separating stems from affixes, we extracted several affixes which are reported in Table \ref{affix_}. We emphasize that there might be wrong segmentations in Morfessor's output, e.g. Turkish is a suffix-based language, so there are no prefixes in this language, but based on what Morfessor generated we extracted $11$ different types of prefixes. We do not post-process Morfessor's outputs.  
\begin{table}[th]
	\centering
	\small
	\begin{tabular}{c  c  c}
	\toprule
	Language & Prefix & Suffix\\\midrule
	German & $75$ & $160$\\
	Russian & $110$ & $260$\\
	Turkish  &$11$ & $293$\\
	\bottomrule
    \end{tabular}
    \caption{\label{affix_}The number of affixes extracted for each language.}
\end{table}

Using the neural language model we train word, stem, and affix embeddings, and initialize the look-up table (but not other parts) of the decoder using those affixes. The look-up table includes high-quality affixes trained on the target side of the parallel corpus by which we train the translation model. Clearly, such an affix table is an additional knowledge source for the decoder. It preserves information which is very close to what the decoder actually needs. However, there might be some missing pieces of information or some incompatibility between the decoder and the table, so we do not freeze the morphology table during training, but let the decoder update it with respect to its needs in the forward and backward passes.

\subsection{Experimental Results}\label{exp2}
Table \ref{final-res} summarizes our experimental results. We report results for the {\it bpe}$\rightarrow${\it char} setting, which means the source token is a {\it bpe} unit and the decoder samples a character at each time step. CDNMT is the baseline model. Table \ref{final-res} includes scores reported from the original CDNMT model \cite{chung-cho-bengio} as well as the scores from our reimplementation. To make our work comparable and show the impact of the new architecture, we tried to replicate CDNMT's results in our experimental setting, we kept everything (parameters, iterations, epochs etc.) unchanged and evaluated the extended model in the same setting. Table \ref{final-res} reports BLEU scores \citep{papineni2002bleu} of our NMT models. 
\begin{table}[th]
	\centering
	\small
	\begin{tabular}{l  c  c  c}
		\toprule
		{\bf Model} & {\bf  En$\rightarrow$De} & {\bf  En$\rightarrow$Ru} & {\bf  En$\rightarrow$Tr}\\\midrule
		CDNMT &        21.33  & 26.00  & - \\
		CDNMT$^{*}$ &       21.01  & 26.23  & 18.01\\
		CDNMT$^{*}_{m}$ & \textbf{21.27}  & \textbf{26.78}  & \textbf{18.44}\\
		CDNMT$^{*}_{o}$ & \textbf{21.39}  & 26.39  & \textbf{18.59} \\
		CDNMT$^{*}_{mo}$& \textbf{21.48}  & \textbf{26.84}  & \textbf{18.70} \\
		\bottomrule
	\end{tabular}
	\caption{\label{final-res} CDNMT$^{*}$ is our implementation of CDNMT. $m$ and $o$ indicates that the base model is extended with the morphology table and the additional output channel, respectively. $mo$ is the combination of both the extensions. The improvement provided by the boldfaced number compared to CDNMT$^{*}$ is statistically significant according to paired bootstrap re-sampling \citep{koehn2004statistical} with $p=0.05$.}
\end{table}	

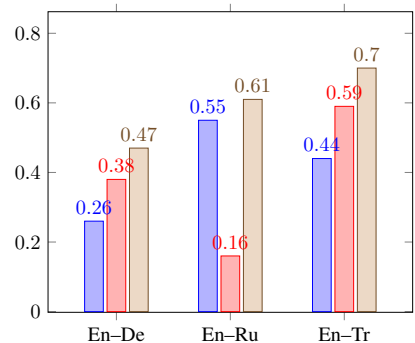
\begin{figure}[h]
\centering
\begin{tikzpicture}[yscale=0.7,xscale=0.7]
\begin{axis}[
    ybar,
    enlargelimits=0.3,
    legend style={at={(0.5,-0.15)},
      anchor=north,legend columns=-1},
    symbolic x coords={En--De,En--Ru,En--Tr},
    xtick=data,
    nodes near coords,
    nodes near coords align={vertical},
    ]
\addplot coordinates {(En--De,0.26) (En--Ru,0.55) (En--Tr,0.44)};
\addplot coordinates {(En--De,0.38) (En--Ru,0.16) (En--Tr,0.59)};
\addplot coordinates {(En--De,0.47) (En--Ru,0.61) (En--Tr,0.70)};
%\legend{\textit{m},\textit{o},\textit{mo}}
\end{axis}
\end{tikzpicture}
\caption{\label{bars}The $y$ axis shows the difference between the BLEU score of CDNMT$^*$ and the extended model. The first, second, and third bars show the $m$, $o$, and $mo$ extensions, respectively.}
\end{figure}

Table \ref{final-res} can be interpreted from different perspectives but the main findings are summarized as follows: 
\begin{itemize}
    \item The morphology table yields significant improvements for all languages and settings.
    \item The morphology table boosts the En--Tr engine more than others and we think this is because of the nature of the language. Turkish is an agglutinative language in which morphemes are clearly separable from each other, but in German and Russian morphological transformations rely more on fusional operations rather than agglutination.
    \item It seems that there is a direct relation between the size of the morphology table and the gain provided for the decoder, because Russian and Turkish have bigger tables and benefit from the table more than German which has fewer affixes. %It seems a bigger morphology table with more affixes provides higher BLEU scores.
    \item The auxiliary output channel is even more useful than the morphology table for all settings but En--Ru, and we think this is because of the morpheme-per-word ratio in Russian. The number of morphemes attached to a Russian word is usually more than those of German and Turkish words in our corpora, and it makes the prediction harder for the classifier (the more the number of suffixes attached to a word, the harder the classification task). 
    
    \item The combination of the morphology table and the extra output channel provides the best result for all languages.
\end{itemize}
Figure \ref{bars} depicts the impact of the morphology table and the extra output channel for each language.
\begin{figure*}[th]
    \centering
       \includegraphics[width=0.98\textwidth]{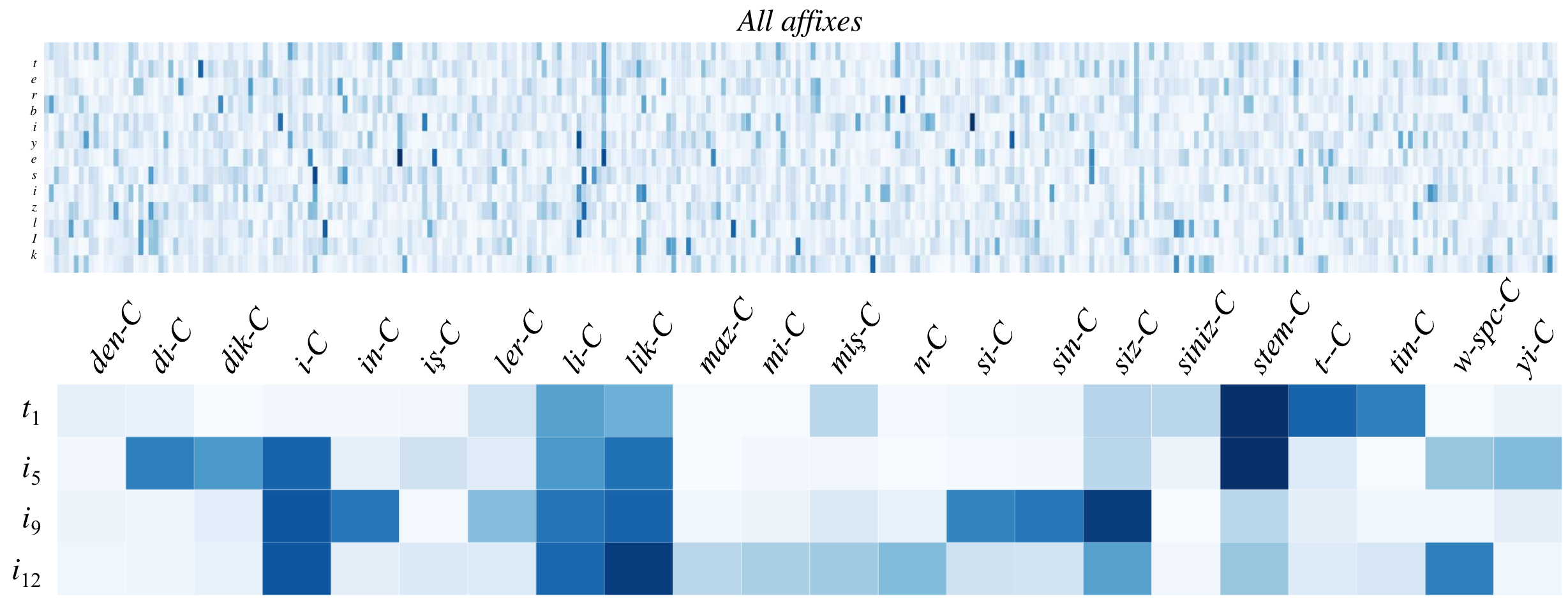}
        %\caption{A gull}
        \label{fig:gull}
    %\begin{subfigure}
    %    \includegraphics[width=0.95\textwidth]{visu-2}
        %\caption{A tiger}
    %    \label{fig:tiger}
    %\end{subfigure}
    \caption{\small\label{vis}Visualizing the attention weights between the morphology table and the decoder when generating `\textit{terbiyesizlik}.}
\end{figure*}

To further study our models' behaviour and ensure that our extensions do not generate random improvements we visualized some attention weights when generating `\textit{terbiyesizlik}'. In Figure \ref{vis}, the upper figure shows attention weights for all Turkish affixes, where the \textit{y} axis shows different time steps and the \textit{x} axis includes attention weights of all affixes (304 columns) for those time steps, e.g. the first row and the first column represents the attention weight assigned to the first Turkish affix when sampling \textit{\textbf{t}} in `\textit{\textbf{t}erbiyesizlik}'. While at the first glance the figure may appear to be somewhat confusing, but it provides some interesting insights which we elaborate next. 

In addition to the whole attention matrix we also visualized a subset of weights to show how the morphology table provides useful information. In the second figure we study the behaviour of the morphology table for the first (\textit{\textbf{t}$_1$}), fifth (\textit{\textbf{i}$_5$}), ninth (\textit{\textbf{i}$_{9}$}), and twelfth (\textit{\textbf{i}$_{12}$}) time steps when generating the same Turkish word `\textit{{\textbf{t}}$_1$erb\textbf{i}$_5$yes\textbf{i}$_9$zl\textbf{i}$_{12}$k}'. \textit{\textbf{t}$_1$} is the first character of the word. We also have three \textit{\textbf{i}} characters from different morphemes, where the first one is part of the stem, the second one belongs to the suffix `\textit{siz}', and the third one to `\textit{lik}'. It is interesting to see how the table reacts to the same character from different parts. For each time step we selected the top-$10$ affixes which have the highest attention weights. The set of top-$10$ affixes can be different for each step, so we made a union of those sets which gives us $22$ affixes. The bottom part of Figure \ref{vis} shows the attention weights for those $22$ affixes at each time step. 

\begin{table*}[h]
	\centering
	\begin{tabular}{l l}
	\textbf{Reference}: & {\it Gesch{\"a}ftsbedingungen f{\"u}r das Senden     von Beitr{\"a}gen an  die BBC}\\
	
	\textbf{CDNMT$^*$} & {\it allgemeinen geschaftsbedingungen fur die versendung von Beitr\"{a}gen an  die BBC}\\
	
	\textbf{CDNMT$^*_{mo}$} & {\it  Gesch{\"a}ft s bedingungen f{\"u}r die versendung von Beitr{\"a}gen zum     BBC}
    \end{tabular}
    \caption{\small\label{trans_}Comparing translation results for the CDNMT$^*$ (baseline) and CDNMT$^*_{mo}$ (improved) models when the input sentence is `{\it Terms and conditions for sending contributions to the BBC}'.}
\end{table*}

After analyzing the weights we observed interesting properties about the morphology table and the auxiliary attention module.\footnote{Our observations are not based on this example alone as we studied other random examples, and the table shows consistent behaviour for all examples.} The main findings about the behaviour of the table are as follows: 
\begin{itemize}
    \item The model assigns high attention weights to \textit{stem-C} for almost all time steps. However, the weights assigned to this class for \textit{\textbf{t}$_1$} and \textit{\textbf{i}$_5$} are much higher than those of affix characters (as they are part of the stem). The vertical lines in both figures approve this feature (bad behaviour).
    
    \item For some unknown reasons there are some affixes which have no direct relation to that particulate time step but they receive a high attention, such as \textit{maz} in \textit{\textbf{t}$_{12}$} (bad behaviour).
    
    \item For almost all time steps the highest attention weight belongs to the class which is expected to be selected, e.g. weights for (\textit{\textbf{i}$_5$},\textit{stem-C}) or (\textit{\textbf{i}$_{9}$},\textit{siz-C}) (good behaviour).
    
    \item The morphology table may send bad or good signals but it is consistent for similar or co-occurring characters, e.g. for the last three time steps \textit{\textbf{l}$_{11}$}, \textit{\textbf{i}$_{12}$}, and \textit{\textbf{k}$_{13}$}, almost the same set of affixes receives the highest attention weights. This consistency is exactly what we are looking for, as it can define a reliable external constraint for the decoder to guide it. Vertical lines on the figure also confirm this fact. They show that for a set of consecutive characters which belong to the same morpheme the attention module sends a signal from a particular affix (good behaviour).
    
    \item There are some affixes which might not be directly related to that time step but receive high attention weights. This is because those affixes either include the same character which the decoder tries to predict (e.g. \textit{i-C} for \textit{\textbf{i}$_{4}$} or \textit{t-C} and \textit{tin-C} for \textit{\textbf{t}$_{1}$}), or frequently appear with that part of the word which includes the target character (e.g. \textit{mi-C} has a high weight when predicting \textit{\textbf{t}$_1$} because \textit{\textbf{t}$_1$} belongs to \textit{terbiye} which frequently collocates with \textit{mi-C}: \textit{terbiye+mi}) (good behaviour).
\end{itemize}

Finally, in order to complete our evaluation study we feed the English-to-German NMT model with the sentence  `{\it Terms and conditions for sending contributions to the BBC}', to show how the model behaves differently and generates a better target sentence. Translations generated by our models are illustrated in Table \ref{trans_}.

The table demonstrates that our architecture is able to control the decoder and limit its selections, e.g. the word {\it `allgemeinen'} generated by the baseline model is redundant. There is no constraint to inform the baseline model that this word should not be generated, whereas our proposed architecture controls the decoder in such situations. After analyzing our model, we realized that there are strong attention weights assigned to the \textit{w-space} (indicating white space characters) and \textit{BOS} (beginning of the sequence) columns of the affix table while sampling the first character of the word {\it `Gesch{\"a}ft'}, which shows that the decoder is informed about the start point of the sequence. Similar to the baseline model's decoder, our decoder can sample any character including \textit{`a'} of {\it `allgemeinen'} or \textit{`G'} of {\it `Gesch{\"a}ft'}. Translation information stored in the baseline decoder is not sufficient for selecting the right character \textit{`G'}, so the decoder wrongly starts with \textit{`i'} and continues along a wrong path up to generating the whole word. However, our decoder's information is accompanied with signals from the affix table which force it to start with a better initial character, whose sampling leads to generating the correct target word. 

Another interesting feature about the table is the new structure `{\it  Gesch{\"a}ft s bedingungen'} generated by the improved model. As the reference translation shows, in the correct form these two structures should be glued together via \textit{`s'}, which can be considered as an infix. As our model is supposed to detect this sort of intra-word relation, it treats the whole structure as two compounds which are connected to one another via an infix. Although this is not a correct translation and it would be trivial to post-edit into the correct output form, it is interesting to see how our mechanism forces the decoder to pay attention to intra-word relations. 

Apart from these two interesting findings, the number of wrong character selections in the baseline model is considerably reduced in the improved model because of our enhanced architecture.  

\section{Conclusion and Future Work}
In this paper we proposed a new architecture to incorporate morphological information into the NMT pipeline. We extended the state-of-the-art NMT model \citep{chung-cho-bengio} with a morphology table. The table could be considered as an external knowledge source which is helpful as it increases the capacity of the model by increasing the number of network parameters. We tried to benefit from this advantage. Moreover, we managed to fill the table with morphological information to further boost the NMT model when translating into MRLs. Apart from the table we also designed an additional output channel which forces the decoder to predict morphological annotations. The error signals coming from the second channel during training inform the decoder with morphological properties of the target language. Experimental results show that our techniques were useful for NMT of MRLs. 

As our future work we follow three main ideas. $i$) We try to find more efficient ways to supply morphological information for both the encoder and decoder. $ii$) We plan to benefit from other types of information such as syntactic and semantic annotations to boost the decoder, as the table is not limited to morphological information alone and can preserve other sorts of information. 
$iii$) Finally, we target sequence generation for fusional languages. Although our model showed significant improvements for both German and Russian, the proposed model is more suitable for generating sequences in agglutinative languages.

\section*{Acknowledgments}
We thank our anonymous reviewers for their valuable feedback, as  well as the Irish centre for high-end computing (\url{www.ichec.ie}) for providing computational infrastructures. This work has been supported by the ADAPT Centre for Digital Content Technology which is funded under the SFI Research Centres Programme (Grant 13/RC/2106) and is co-funded under the European Regional Development Fund.
\bibliography{naaclhlt2018}
\bibliographystyle{acl_natbib}
\end{document}